\documentclass{isprs} % isprs class modified 23-04-2019 (Dennis Wittich)
\usepackage{subfigure}
\usepackage{setspace}
\usepackage{geometry} % added 27-02-2014 Markus Englich
\usepackage{epstopdf}
\usepackage[labelsep=period]{caption}  % added 14-04-2016 Markus Englich - Recommendation by Sebastian Brocks
\usepackage[british]{babel} 
\usepackage[hang]{footmisc}
\usepackage{algorithm2e}
\usepackage{comment}
\usepackage{url}
\usepackage{tabularx}
\usepackage{xltabular}
\usepackage{multirow}
\usepackage{booktabs}
\usepackage{graphicx}
\usepackage{subfig}
\usepackage{amsmath}
\usepackage{amssymb}
\usepackage{xcolor}

\begin{document}

\title{TOWARDS ACCURATE INSTANCE SEGMENTATION IN LARGE-SCALE\\LIDAR POINT CLOUDS}
\date{}

% KAO: Remove extra spacing
% Anonymous submissions, authors' names should not be visible
\author{Binbin Xiang\textsuperscript{1}, Torben Peters\textsuperscript{1}, Theodora Kontogianni\textsuperscript{1}, Frawa Vetterli\textsuperscript{1}, Stefano Puliti\textsuperscript{2}, Rasmus Astrup\textsuperscript{2}, Konrad Schindler\textsuperscript{1}}% (for review, names must be rendered anonymous)}

% \thanks{Corresponding author} 

% KAO: Remove extra newline
% Anonymous submissions, authors' affiliations should not be visible
\address{\textsuperscript{1}ETH Zürich, Switzerland - (bxiang, tpeters, tkontogianni, vfrawa, schindler)@ethz.ch\\ \textsuperscript{2}Norwegian Institute of Bioeconomy Research (NIBIO) - (Stefano.Puliti, rasmus.astrup)@nibio.no}% (for review, affiliations must be rendered anonymous)}

% If the corresponding author is NOT the final author, always add a % space before the subsequent comma, i.e.
% first author name\textsuperscript{a,}\thanks{Corresponding author} , % second author name \textsuperscript{b}, etc.
% thanks to Niclas Borlin 05-05-2016

\commission{XX, }{YY} %This field is optional. If filled, XX and YY should be replaced by adequate numbers. See https://www2.isprs.org/commissions/
\workinggroup{XX/YY} %This field is optional.
\icwg{}   %This field is optional.

% KAO: Use times symbol
%100--250 words
\abstract{
Panoptic segmentation is the combination of semantic and instance segmentation: assign the points in a 3D point cloud to semantic categories \emph{and} partition them into distinct object instances. It has many obvious applications for outdoor scene understanding, from city mapping to forest management. Existing methods struggle to segment nearby instances of the same semantic category, like adjacent pieces of street furniture or neighbouring trees, which limits their usability for inventory- or management-type applications that rely on object instances. This study explores the steps of the panoptic segmentation pipeline concerned with clustering points into object instances, with the goal to alleviate that bottleneck. We find that a carefully designed clustering strategy, which leverages multiple types of learned point embeddings, significantly improves instance segmentation. Experiments on the \emph{NPM3D} urban mobile mapping dataset and the \emph{FOR-instance} forest dataset demonstrate the effectiveness and versatility of the proposed strategy.
}

\maketitle

%\saythanks % added 28-02-2014 Markus Englich

\section{INTRODUCTION}\label{INTRODUCTION}
 
% KAO: Sloppy spacing ensures non-overfull lines. Can be removed if this is not an issue.
\sloppy

Laser scanning has emerged as a main sensing technology to digitise 3D scenes, thanks to its ability to deliver dense 3D point observations with high reliability. The unstructured point clouds it produces are, however, not directly usable as a product (except for visualisation) and must be processed further to extract meaningful entities for mapping and analysis. Panoptic segmentation~\cite{Kirillov2019,Zhou2021} addresses the case where the desired entities are semantically meaningful objects, like individual trees or traffic signs.
\footnote{As opposed to low-level primitives without semantic meaning, such as salient keypoints or planar surfaces.}

Panoptic segmentation is a generic and versatile processing step that may be useful across many different fields. In the context of street scenes, it facilitates scene understanding and mapping at the level of objects, like buildings, traffic signs, pedestrians, etc.~\cite{Fong2022,Chen2022}, which in turn supports applications from urban planning to autonomous vehicles. In forest regions, panoptic segmentation can localise and delimit individual trees, which in turn supports applications like resource management, environmental protection and ecological restoration~\cite{Calders2020}.

Large-scale outdoor point clouds pose particular challenges for panoptic segmentation. Besides common problems of point cloud processing, such as occlusions, moving objects and a large range of object scales and point densities~\cite{Chen2016Dynamic}, an important issue is the lack of natural ``processing units'': unlike indoor scans that can be processed on a per-room basis~\cite{Armeni2016,Dai2017ScanNet} or panoramic scans from robotic systems that can be processed on a per-scan basis~\cite{Behley2019SemanticKITTI}, there is no natural way of dividing an outdoor scene into independent subsets.
A specific difficulty of panoptic segmentation is the requirement to separate objects of the same category, which can be considerably harder than only assigning semantic labels to points, as in the case of trees with overlapping crowns. 

Modern panoptic segmentation techniques are often built upon a 3D deep network backbone that extracts per-point features, followed by network branches that segment the points into semantic categories and into object instances, based on those features. The backbone network is not the focus of this paper. We treat it as a plug-in module of our overall network that ingests a point cloud and returns a feature vector of fixed length for every point. Multiple well-proven, trainable feature extractors exist for the task~\cite{Thomas2019KPConv,Choy20194D}. Semantic segmentation also has reached a certain level of maturity and can be regarded as a commodity. Technically, the associated network branch is a classifier that maps the feature representation to a list of (pseudo-)probabilities per point and is typically trained by minimising the cross-entropy loss. We follow that practice, but do not deeply delve into the details. The focus of the present paper is on the instance segmentation branch, arguably the least explored part of the problem and the current performance bottleneck. There are two different strategies to identify object instances in point clouds.

\subsection{Top-down instance detection}\label{sec:Intro_Top-down}

% KAO: Remove spacing before label: can cause reference to be wrong
\begin{figure*}[!t]
\centering
\includegraphics[width=0.985\textwidth]{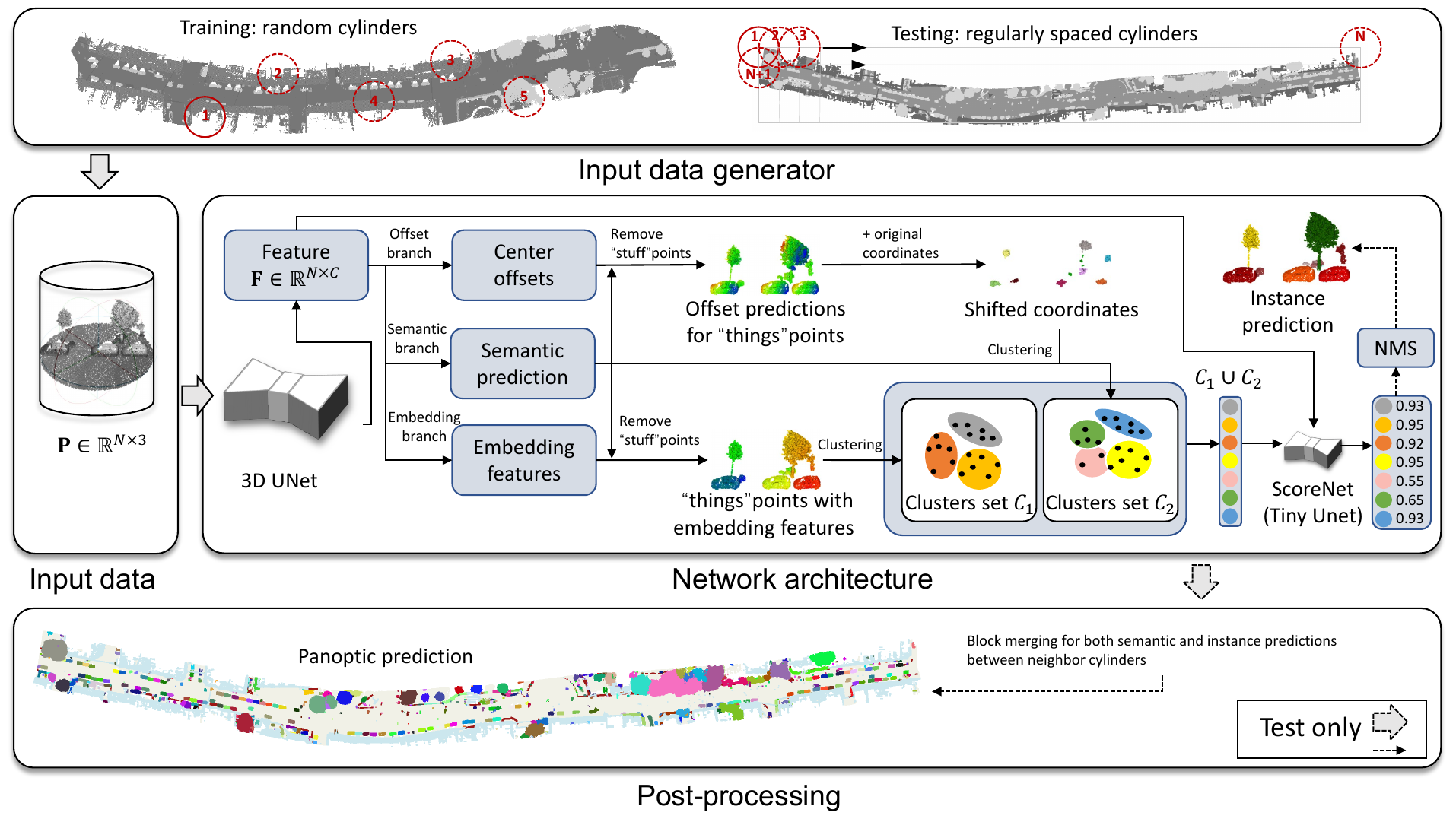}
 \vspace{-1em}
 \caption{The bottom-up panoptic segmentation pipeline studied in this paper.}
 %\vspace{0.5em}
\label{fig:pipeline}
\end{figure*}

The top-down approach first performs object detection to obtain a set of bounding boxes around 3D object candidates. Then the points inside each box are separated into points on the object and points on the background with a binary classifier. The quality of such methods largely depends on the object detection step. The earliest attempts at instance segmentation were top-down methods~\cite{Yang2019Learning}, following the success of Mask-RCNN~\cite{He2017maskRCNN} in the image domain, but later they were surpassed by bottom-up methods (see below). There is recent evidence that in certain types of (indoor) scenarios the top-down approach is competitive or even superior~\cite{Kolodiazhnyi2023}. Here, we do not further investigate the top-down strategy for two reasons: (1)~For outdoor scenes, bounding box detectors tend to work well only for a small number of categories, especially pedestrians and vehicles~\cite{Zhang2020Instance}, whereas they often miss small objects like bollards, and objects that have greatly varying shape and aspect ratio, for instance trees. (2)~Outdoor mapping point clouds cannot be split into natural entities like rooms, instead they have to be subdivided into arbitrary, computationally manageable chunks using a sliding window or random sampling. Consequently, many objects -- especially large ones -- are cut into parts and only partially visible in each chunk, making them hard to find for an  object detector based on global shape and layout.

\subsection{Bottom-up instance grouping}\label{sec:Intro_Bottom-up}

The bottom-up strategy aims to equip each individual point with an instance-sensitive feature representation, such that instances can be found by clustering the points in the associated feature space. These (learned) instance features are computed with the help of a neural network, based on the point coordinates and/or the backbone features~\cite{Han2020OccuSeg,Lahoud20193D,Engelmann20203DMPA}. A natural feature to find instances is the offset from the point to the instance centre, in the spirit of the (generalised) Hough transform. An important finding in  this context was that the unsupervised clustering step is, by itself, rather unreliable. To address the issue, PointGroup~\cite{Jiang2020PointGroup} proposed to run multiple clustering variants and obtain a redundant set of clusters. The quality of these instance candidates is then estimated with a (learned) ScoreNet, such that they can be sorted by their scores and pruned to an optimal set of instances with non-maximum suppression (NMS). The principle to let multiple candidate segmentations compete, and to thereby benefit from the complementary strengths of different clustering methods, proved to work very well and sparked a series of follow-up works that further explored the idea. MaskGroup~\cite{Zhong2022MaskGroup} clusters at different spatial scales, and SoftGroup~\cite{Vu2022SoftGroup} keeps soft semantic labels, so as to enable clustering across different categories and rectify semantic segmentation errors. HAIS~\cite{Chen2021Hierarchical} adds a MaskNet after the clustering step, which examines each individual cluster and aims to detect and remove points that do not belong to the object instance. \cite{Engelmann20203DMPA} is also based on instance proposals, and refines them by modelling their relations with a graph neural network, which is again more suitable for complete, self-contained target areas like indoor rooms. \cite{Liang2021Instance} construct a cluster hierarchy and traverse the associated tree to generate proposals, which are then again assessed with a ScoreNet. For the present study we also build on the PointGroup principle. We note that, contrary to most other existing work, it has also been shown to work in outdoor settings.

For completeness, we mention that directly clustering points into instances in 3D scene space -- arguably the most obvious strategy -- does not work well for many mapping tasks. For compact, well-separated objects like vehicles on the road this strategy can work quite well~\cite{Zhao2021Technical}, but it tends to fail as soon as objects are located in close proximity, such as tightly parked cars; or even touch, like the crowns of nearby trees. There is a consensus in the literature that in such situations a-priori knowledge about the objects' shapes and layouts is required, e.g.,~\cite{Lahoud20193D,Engelmann20203DMPA,Jiang2020PointGroup}. Conveniently, that knowledge is also needed for semantic segmentation, so it can be derived from the same latent features with little computational overhead.

Recently, transformer-type neural networks were applied for instance segmentation~\cite{Liu20223DQueryIS,Schult2023Mask3D,Sun2022Superpoint}, following a trend in the 2D image domain~\cite{Zhang2021KNet,Cheng2022Masked}. The principle is to replace the explicit instance feature extraction and clustering step with instance queries, based on the attention mechanism of the transformer architecture. So far these methods have only been demonstrated on indoor datasets.  They appear to be particularly successful in terms of \emph{detecting} instances, whereas the per-point segmentation performance is on par with PointGroup-style methods. In practical terms, the learned proposal generator is rather elegant, but comes at the cost of significantly higher memory demand. When processing densely scanned outdoor point clouds, GPU memory is the limiting factor even for conventional, convolution-based methods. Hence, we exclude transformers from this study, but note that adapting them to outdoor mapping is an interesting future direction.

\subsection{Contributions}

We have developed an effective deep learning-based workflow for panoptic segmentation of large outdoor mapping point clouds, based on the bottom-up grouping strategy. Design choices for instance segmentation are carefully evaluated and analysed in a series of experiments on two different data sets, one showing streets scenes (NPM3D) and one dense forests (FOR-instance). Our main findings are:
(1)~The often used grouping based on centroid offsets struggles to separate object instances located close to each other. A learned feature embedding, trained with a contrastive loss to discriminate instances, can often separate such instances.
(2)~On the other hand, offset vectors more accurately separate objects that have similar local shape, but are located far from each other. We find that the best results are achieved by combining both methods and letting the subsequent ScoreNet select from both proposal sets.
(3)~Clustering based on embedding features, which does not depend directly on the semantic segmentation result, reduce mistakes caused by incorrect semantic labels and thereby improve the completeness of the affected instances.
(4)~A simple block merging strategy is sufficient to combine the segmentations of local subsets into a coherent large-scale panoptic segmentation map.
(5)~State-of-the-art methods for 3D panoptic segmentation work well even for challenging tasks like separating tree crowns in dense forests. We expect those methods to be more widely adopted for practical applications in the near future.

\section{METHOD}\label{sec:Methodology}

The overall pipeline of our proposed method, shown in Figure~\ref{fig:pipeline}, consists of three main components: an input data generator (Section~\ref{sec:InputDataGenerator}), a deep neural network (Section~\ref{sec:NetworkArchitecture}), and a post-processor (Section~\ref{sec:Post-processing}).

\subsection{Input data generator}\label{sec:InputDataGenerator}
\vspace{-0.25em}

As a first step, the entire point cloud is voxel-grid subsampled to sparsify overly dense regions and achieve a homogeneous (maximum) point density. The voxel size for the filter depends on the scene. In our implementation we use 12$\times$12$\times$12$\,$cm\textsuperscript{3} (579~points/m\textsuperscript{3}) for urban scenes and 20$\times$20$\times$20$\,$cm\textsuperscript{3} (125~points/m\textsuperscript{3}) for forest scenes, see Table~\ref{table:parameterSettings}. These values were chosen based on extensive ablation studies performed in~\cite{Xiang2023Review}. Even so, outdoor scans are far too large to process as a whole on existing hardware. As an example, a single scene from the NPM3D dataset~\cite{Roynard2018ParisLille}, covering a stretch of road of length $\approx\,$600$\,$m, has several million points. It is therefore necessary to process the data in local blocks. When applying the trained network to new data these blocks can be sampled in sliding-window fashion. During training, we simply sample them randomly. There are different ways to define the local neighborhood of points that constitutes a block around a sampled location, popular choices include cubic boxes, spheres or cylinders. We opt for the cylinder, for the following reasons: (1) It avoids cutting objects along the vertical, which on the one hand improves the handling of long vertical objects such as street lamps or trees, and on the other hand simplifies block merging to a 2D problem (Section~\ref{sec:Post-processing}). (2) It ensures computational efficiency. When working with large point clouds, the computational bottleneck is not error back-propagation, but rather geometric queries like finding neighbours. Cylindrical neighbourhoods are compliant with efficient algorithms like fast radius search (normally implemented via spatial search structures such as KD-trees, and available in the Torch-Point3D framework~\cite{Chaton2020TorchPoints3D}).

To sample training cylinders in a way that ensures sufficient coverage of rare categories, we take inspiration from KPConv~\cite{Thomas2019KPConv}: the location of the vertical cylinder axis is found by randomly sampling one of the training data points, with  sampling probabilities proportional to the square root of the inverse class frequencies, $P_i\propto\sqrt{1/N_i}$. After sampling a fixed training set of many cylindrical blocks, the points' $(x,y)$-coordinates in each of them are shifted to have their origin in the cylinder centre. Moreover, various data augmentation techniques are applied: isotropic, additive Gaussian random noise on the point coordinates (jittering), random rotations around the cylinder axis, random anisotropic scaling by factors $s\in[0.9,1.1]$, and random reflection along the $y$-axis. At test time the cylindrical blocks are sampled regularly with fixed step size along the $(x,y)$-grid so as to ensure even coverage of the point cloud, see the illustration of the input data generator in Figure~\ref{fig:pipeline}.

\subsection{Network architecture}\label{sec:NetworkArchitecture}
\vspace{-0.25em}

As {\bf feature extraction backbone} we use the Minkowski Engine~\cite{Choy20194D}, which offers a favourable trade-off between performance and computational cost~\cite{Xiang2023Review}. In a nutshell, it is a 3D U-Net that operates on the voxelised point cloud with sub-manifold sparse convolutions. The resulting per-point feature vectors of length 16 serve as input for three output branches: one that estimates point-wise semantic labels, one that regresses offsets to the instance center, and one that extracts instance-discriminative embedding features.

\vspace{-0.25em}
The \textbf{semantic segmentation branch} consists only of a multi-layer perceptron (MLP) with a single hidden layer with \emph{softmax} activations and outputs semantic class probabilities for each point. That branch is trained with a standard cross-entropy loss. Semantic labels are obtained by taking the \emph{argmax} over the predicted category probabilities. Points assigned to categories that cannot be divided into well-defined instances (so-called ``stuff'' categories, like for instance ``road'' or ``building facade'') are ignored during instance segmentation.

\vspace{-0.25em}
The \textbf{centre offset branch}, advocated by several studies about instance segmentation~\cite{Jiang2020PointGroup,Vu2022SoftGroup,Zhong2022MaskGroup,Chen2021Hierarchical}, operates in 3D scene space: it takes as input the latent encoding extracted by the backbone and, for each point, predicts a 3D offset vector that would take that point to the estimated instance centre. I.e., if the predictions were perfect then shifting all points by their offsets would collapse each instance to a single point. The corresponding loss function is a combination of (1) the cosine distance between the true and predicted offset vectors and (2) the $L_1$ distance between their endpoints.

\makeatletter
\setlength{\smallskipamount}{1pt}
\makeatother

\RestyleAlgo{ruled}
%% This is needed if you want to add comments in
%% your algorithm with \Comment
\SetKwComment{Comment}{/* }{ */}
\begin{algorithm}[tb]
%\small
\caption{Block Merging}\label{alg:CylinderMerging}
\SetKwInOut{Input}{Input}
\SetKwInOut{Output}{Output}
\Input{%
- list of blocks $B_i$\\
- list of point indices $I_{i,j}$ for every instance $j$\\
$\,\,\,\,$in each block $i$\\
- overlap threshold $T_\text{IoU}$
}
\smallskip
\smallskip
\Output{%
- global per-point label vector $P$
}
\vspace{4pt}
initialise all elements of $P$ to $-1$\;
\smallskip
set instance counter $q \gets 1$\;
\smallskip
\For{\normalfont every block $B_i$}{
  \smallskip
  \For{\normalfont every instance $I_{i,j}$ in $B_i$}{
     \smallskip
     \uIf{\normalfont all $P(I_{i,j})=-1$}{
        \smallskip
        set all $P(I_{i,j}) \gets q$\;
        $q \gets q+1$\;
     }
     \smallskip
     \uElseIf{\normalfont all $P(I_{i,j})\neq-1$}{
        \smallskip
        continue\;
     }
     \Else{
       $J_r \gets$ instance in $P$ with highest IoU to $I_{i,j}$\;
       \smallskip
       $r \gets$ instance label of $J_r$\;
       \smallskip
       \uIf{\normalfont $\text{IoU}(J_r,I_{i,j})>T_\text{IoU}$}{
    \smallskip
          $P(I_{i,j}=-1) \gets r$\;
       }
       \Else{
          $P(I_{i,j}=-1) \gets q$\;
          $q \gets q+1$\;
     }
   }
 }
}
\end{algorithm}

The \textbf{instance embedding branch} also ingests the latent encoding from the backbone. Instead of trying to find the geometric object centre, it embeds each point in a 5D feature space that is optimised to discriminate between instances. The embedding is supervised with a contrastive loss function that favours small distances between points from the same instance and large distances between points from different instances. Importantly, the embedding space has more than three dimensions, hence it has some spare capacity to represent object properties beyond being a compact cluster around a 3D centre point.

We found that the two ways of measuring point-to-instance affinities, either by regressing explicit centre offsets in geometric space~\cite{Jiang2020PointGroup} or by contrastive embedding~\cite{DeBrabandere2017,Wang2019Associatively}, complement each other. In fact, instance segmentation based on local 3D point configurations must balance different a-priori expectations. On the one hand, points that form a compact structure surrounded by empty space are indeed likely to belong to the same object, and that situation is easy to encode in the form of centroid offsets -- e.g., for a local region on an isolated car one can often guess the direction to the object centre just from the local surface shape. On the other hand, when objects are located near each other it becomes important to look past proximity -- e.g., for a region of a forest canopy it is often easy to say which tree it belongs to, but nevertheless difficult to point to a clear object centre. This is why we employ both strategies.

The predicted offsets are simply applied to the 3D point coordinates to shift them to the estimated object centre, and then clustered into instance candidates by region growing with a distance threshold. Note, mapping the latent features to 3D offset vectors discards the semantic category information originally contained in the features. Hence, the clustering is constrained to only include points from the same category in a candidate instance. In the 5D embedding space, where distances do not have a direct geometric meaning, candidates are found with mean-shift clustering.

From the redundant set of instance candidates, we want to retain the subset that best explains the scene. To that end we train a network branch to predict how well each candidate matches a ground truth instance. This ScoreNet regresses the highest expected IoU between the candidate and any of the actual objects. It is a small 3D U-Net model on top of the backbone features, followed by max-pooling and a fully connected layer that outputs a scalar score between 0 and 1 per candidate.

\subsection{Post-processing}\label{sec:Post-processing}

After scoring we are left with an over-complete list of instance candidates, each consisting of a subset of the 3D point cloud, and equipped with an estimate of its goodness-of-fit to some actual object instance. These are post-processed into a final set of instances in the following way. First, clusters with very few points (in our implementation, \textless10; see Table~\ref{table:parameterSettings} for a complete list of hyper-parameters) are discarded. Second, we perform non-maximum suppression (NMS) based on the predicted scores to get rid of redundant clusters. Third, clusters with low scores are also discarded. Having obtained our final estimate per cylindrical block, we run block merging to combine them into a single result for the entire region of interest, see Algorithm~\ref{alg:CylinderMerging}. In brief, the block merging re-assigns instance IDs such that they are globally unique, and greedily fuses instances that were split between different blocks.

After block merging we have a final segmentation of the voxel-grid subsampled point cloud into semantic categories and into object instances. As a final step, we upsample all labels back from the voxel-gridded point cloud to the complete, original one with the nearest-neighbour method. Instance labels for ``stuff'' classes that do not have well-defined instances are set to $-1$. 

\section{EXPERIMENTS}\label{sec:Experiments}

\begin{table}[bt]
%\centering
\resizebox{\columnwidth}{!}{%
\begin{tabular}{lcc}
\toprule
Parameter & NPM3D     & FOR-instance  \\ \midrule
%Iterations for training                                  & 300k      & 450k          \\
Base learning rate of \emph{Adam} optimizer   & 0.001 & 0.001\\
Batch size                              & 4 & 4\\
Voxel side length (m)                          & 0.12 & 0.2\\
Cylinder radius (m)                     & 16 & 4\\
Region growing radius (m)               & 0.03 & 0.03\\
Mean-shift bandwidth\textsuperscript{\tiny$\diamondsuit$} & 0.6 & 0.6\\
Minimum cluster size (\#points)         & 10 & 10\\
Score threshold for discarding clusters & 0.6 & 0.5\\
IoU threshold during NMS                & 0.3 & 0.3\\
IoU threshold during block merging      & 0.01 & 0.01\\ \bottomrule
\multicolumn{3}{l}{\scriptsize $^{\diamondsuit^{\textcolor{white}{\diamondsuit}}}\!\!\!\!$Bandwidth is relative to the specified cluster radius of 1 unit in embedding space}
\end{tabular}
}
\vspace{-0.5em}
\caption{Default parameter settings.}
\label{table:parameterSettings}
\end{table}

\subsection{Experimental settings}\label{sec:ExperimentalSettings}

\noindent\textbf{Datasets.} For our experiments we use two datasets, \emph{NPM3D}~\cite{Roynard2018ParisLille} and \emph{FOR-instance}.
NPM3D consists of mobile laser scanning (MLS) point clouds collected in four different regions in the French cities of Paris and Lille, where each point has been annotated with two labels: one that assigns it to one out of 10 semantic categories and another one that assigns it to an object instance. When inspecting the data, we found 9 cases where multiple tree instances had not been separated correctly (i.e., they had the same ground truth instance label). These cases were manually corrected using the CloudCompare software (https://www.cloudcompare.org, last accessed 03/2023), and 35 individual tree instances were obtained. Our variant of the dataset with 10 semantic categories and enhanced instance labels is publicly available.%
\footnote{https://doi.org/10.5281/zenodo.8118986}

The FOR-instance dataset is a recent benchmark dataset from the forestry domain, aimed at tree instance segmentation and biophysical parameter retrieval. The point clouds were collected from drones equipped with survey-grade laser scanners such as the Riegl VUX-1 UAV and Mini-VUX. The dataset covers diverse regions and forest types across multiple countries. For our purposes, we removed points assigned to the category ``outpoints'' (i.e., partially observed tree instances on region borders), leaving us with only two semantic categories, \emph{tree} and \emph{non-tree} (where the latter includes the forest floor). The panoptic segmentation task thus becomes to separate trees from non-trees and to divide the tree class into individual instances.

Both datasets have been released only recently. Previous outdoor point cloud datasets either did not provide instance annotations or were too small to train deep neural networks, consequently there are hardly any baseline results to compare to. We have made both the data and our source code\footnote{https://github.com/bxiang233/PanopticSegForLargeScalePointCloud} publicly available for future reference.

\noindent\textbf{Evaluation Metrics.} Semantic segmentation quality is measured by the mean intersection-over-union (mIoU) across all categories. To assess instance segmentation we follow~\cite{Wang2019Associatively} and compute the mean precision (mPrec) and mean recall (mRec) over all instances, the corresponding F1-score (harmonic mean of precision and recall), as well as the mean coverage (mCov), defined as the average IoU between ground truth instances and their best-matching instance predictions. We also calculate a variant of mCov that weights instances by their ground truth point count (mWCov). For the combined panoptic segmentation quality we adopt the metrics proposed by~\cite{Kirillov2019}, segmentation quality (SQ), recognition quality (RQ) and panoptic quality (PQ).

\noindent\textbf{Implementation Details.} Our source code is based on the Torch-Point3D library~\cite{Chaton2020TorchPoints3D}. Unless explicitly specified for a given experiment, we use the default parameter values listed in Table~\ref{table:parameterSettings}. All experiments were conducted on a machine with 8-core Intel CPU, 8$\,$GB of memory per core, and one Nvidia Titan RTX GPU with 24$\,$GB of on-board memory.

\subsection{Ablation studies on NPM3D}\label{sec:AblationsNPM3D}

Experiments were conducted on NPM3D to investigate the effects of different hyper-parameters. In all ablation studies, the training portions of Lille1\_1, Lille1\_2, and Lille2 serve as training set and the test portion of Paris serves as test set.

\begin{figure*}[!t]
\includegraphics[width=1.0\textwidth,trim={0 3mm 0 0},clip]{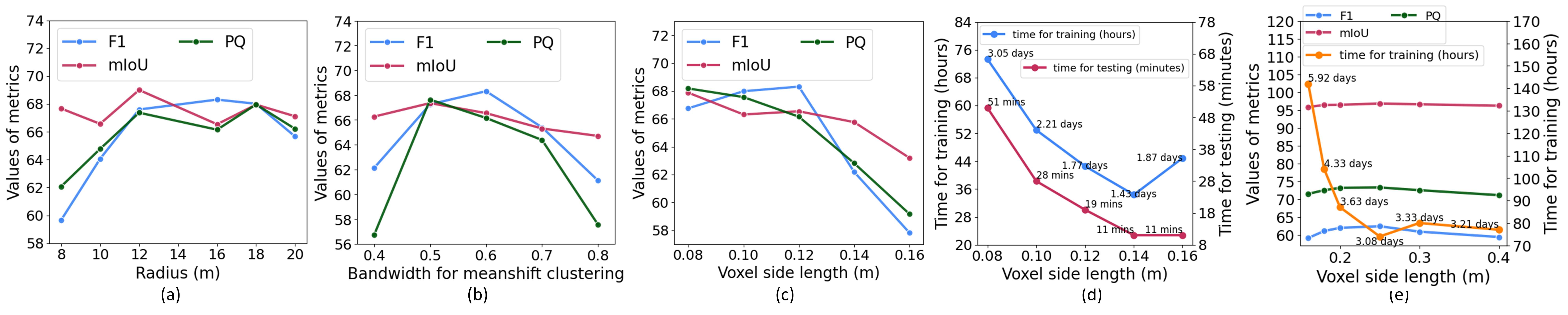}
\vspace{-1.5em}
\caption{Ablation studies. Plots (a)-(d) refer to NPM3D, (e) refers to FOR-instance.}
\vspace{0.5em}
\label{fig:NPM3DAblation}
\end{figure*}

\noindent\textbf{Radius of cylindrical blocks.} As explained, we sample local cylindrical regions from the data to keep computations tractable. A larger cylinder radius means more points, and thus more spatial context, and at the same time fewer incomplete objects and boundary effects; but also slower training and inference. Figures~\ref{fig:NPM3DAblation}a illustrates the impact of the radius on instance segmentation and on semantic segmentation. 
As a general conclusion, the cylinder radius has little influence on the semantic segmentation quality (in terms of mIoU), seemingly a limited amount of local context is sufficient to categorise points. On the contrary, too small blocks markedly degrade instance segmentation (measured by the F1-score), confirming the intuition that it relies more on a complete view of object shape and layout. The performance metrics in Figure~\ref{fig:NPM3DAblation} refer to end-to-end segmentation performance from a system user view, after cylinder merging and upsampling to the original input point cloud. We point out that increasing the cylinder radius from 8$\,$m to 20$\,$m doubles the inference time for the complete set from 12$\,$min to 24$\,$min, and also the training takes roughly twice as long.

\noindent\textbf{Mean-shift bandwidth.} The discriminative training uses the two margins 0.5 and 1.5, meaning that in theory it should bring the feature vectors of all points on an instance to within 0.5 units of the associated cluster centre, whereas there should be a distance of at least 2$\times$1.5 units between two cluster centres~\cite{DeBrabandere2017}. Based on these values we empirically determine the optimal bandwidth of the flat (rectangular) mean-shift kernel, see Figure~\ref{fig:NPM3DAblation}b. Indeed, semantic segmentation performance and the closely related panoptic quality peak at a bandwidth of 0.5, whereas instance segmentation peaks at a slightly higher value of 0.6. 

\noindent\textbf{Voxel grid resolution.} As expected, the overall trend is that point cloud analysis deteriorates with increasing voxel size (stronger down-sampling). As can be seen in Figure~\ref{fig:NPM3DAblation}c, instance segmentation does not benefit from overly dense sampling and reaches its best performance at a voxel size of 12$\times$12$\times$12$\,$cm\textsuperscript{3}. Obviously, smaller voxels significantly increase the computational cost of both training and testing, Figure~\ref{fig:NPM3DAblation}d. We note that the trade-off between resolution, cylinder radius and computational cost depends on the scene properties, c.f.~\ref{fig:NPM3DAblation}e, which is why we chose different values for NPM3D and FOR-instance (Table~\ref{table:parameterSettings}).

\begin{table}[tb]
%\centering
\resizebox{\columnwidth}{!}{%
\begin{tabular}{c|ccc|c}
\toprule
\multicolumn{1}{c|}{\multirow{3}{*}{\begin{tabular}[c]{@{}c@{}} Setting\\ID\end{tabular}}} &
  \multicolumn{3}{c|}{Instance generator} &
  \multirow{3}{*}{\begin{tabular}[c]{@{}c@{}}Use\\ ScoreNet\end{tabular}} \\ 
 \cmidrule{2-4}
\multicolumn{1}{c|}{} &
  \begin{tabular}[c]{@{}c@{}}Raw 3D coords.\\ + region growing\end{tabular} &
  \begin{tabular}[c]{@{}c@{}}Shifted 3D coords.\\ + region growing\end{tabular}
   & \begin{tabular}[c]{@{}c@{}}5D embedding\\+ meanshift\end{tabular} &
   \\ \midrule
{\bf I} & & & \checkmark &\\
{\bf II} & & \checkmark & &\\
{\bf III} & \checkmark & \checkmark & & \checkmark \\
{\bf IV} &  & \checkmark & \checkmark & \checkmark \\
{\bf V} & \checkmark & \checkmark & \checkmark & \checkmark \\ \bottomrule
\end{tabular}%
}
\vspace{-0.5em}
\caption{Summary of the tested instance segmentation settings.}
\label{table:5SettingsforInstanceSeg}
\end{table}

%%%

\begin{comment}
\begin{table}[tb]
%\centering
\resizebox{\columnwidth}{!}{%
\begin{tabular}{c|ccc|cc|c}
\toprule
\multicolumn{1}{c|}{\multirow{3}{*}{Setting ID}} &
  \multicolumn{3}{c|}{Input features} &
  \multicolumn{2}{c|}{Clustering strategy} &
  \multirow{3}{*}{\begin{tabular}[c]{@{}c@{}}Use\\ ScoreNet\end{tabular}} \\ 
 \cmidrule{2-6}
\multicolumn{1}{c|}{} &
  \begin{tabular}[c]{@{}c@{}}Embedding\\ features\end{tabular} &
  \begin{tabular}[c]{@{}c@{}}Original\\ coordinates\end{tabular} &
  \begin{tabular}[c]{@{}c@{}}Shifted\\ coordinates\end{tabular}
   &
  \begin{tabular}[c]{@{}c@{}}Mean-\\ shift\end{tabular} &
  \begin{tabular}[c]{@{}c@{}}Region\\ grow\end{tabular} &
   \\ \midrule
{\bf I} & \checkmark & & & \checkmark & &\\
{\bf II} & & & \checkmark &  & \checkmark &\\
{\bf III} & & \checkmark & \checkmark &  & \checkmark & \checkmark \\
{\bf IV} & \checkmark & & \checkmark &  \checkmark & \checkmark & \checkmark \\
{\bf V} & \checkmark & \checkmark & \checkmark & \checkmark & \checkmark & \checkmark \\ \bottomrule
\end{tabular}%
}
\vspace{-0.5em}
\caption{Summary of the tested instance segmentation settings.}
\label{table:5SettingsforInstanceSeg}
\end{table}
\end{comment}

\subsection{Panoptic segmentation results for NPM3D}
\label{sec:EvaluationOnNPM3D}

\begin{table*}[tb]
\centering
%\resizebox{\textwidth}{!}{%
\begin{tabular}{l|ccccc|c|ccc}
\toprule
\multirow{2}{*}{Setting ID} &
  \multicolumn{5}{c|}{Instance segmentation} &
  \multicolumn{1}{c|}{Semantic seg.} &
  \multicolumn{3}{c}{Panoptic seg.} \\ \cmidrule{2-10} 
 & mCov & mWCov & mPrec & mRec & F1 & mIoU & RQ & SQ & PQ \\ \midrule
{\bf I} (5d embed) & \underline{65.2} & \underline{68.6} & 61.5 & \underline{73.7} & 67.1 & 75.0 & 77.9 & \underline{87.6} & 68.1 \\
{\bf II} (3d offset) & 62.5 & 65.9 & 40.5 & 71.8 & 51.8 & 73.0 & 66.9 & 85.5 & 56.9  \\
{\bf III} (3d offset + 3d raw) & 59.6 & 63.1 & \underline{75.3} & 68.6 & 71.8 & 72.8 & 80.8 & 86.6 & 69.9  \\
{\bf IV} (5d embed + 3d offset) & 63.8 & 67.4 & 74.9 & 73.6 & \textbf{74.3} &\textbf{75.8} & \underline{82.6} & 87.5 & \textbf{72.1}  \\
{\bf V} (5d embed + 3d offset + 3d raw) & 63.2 & 66.7 & 74.0 & 72.7 & 73.3 & 73.8 & 82.1 & 87.2 & 71.4  \\ \bottomrule
\end{tabular}%
%}
\vspace{-0.5em}
\caption{Quantitative results on NPM3D for the five settings listed in Table~\ref{table:5SettingsforInstanceSeg}. All values are percentages [\%].}
\vspace{0.5em}
\label{table:5SettingsResultsNpm3d}
\end{table*}

\begin{figure*}[!t]
\includegraphics[width=1.0\textwidth,trim={0 3mm 0 0},clip]{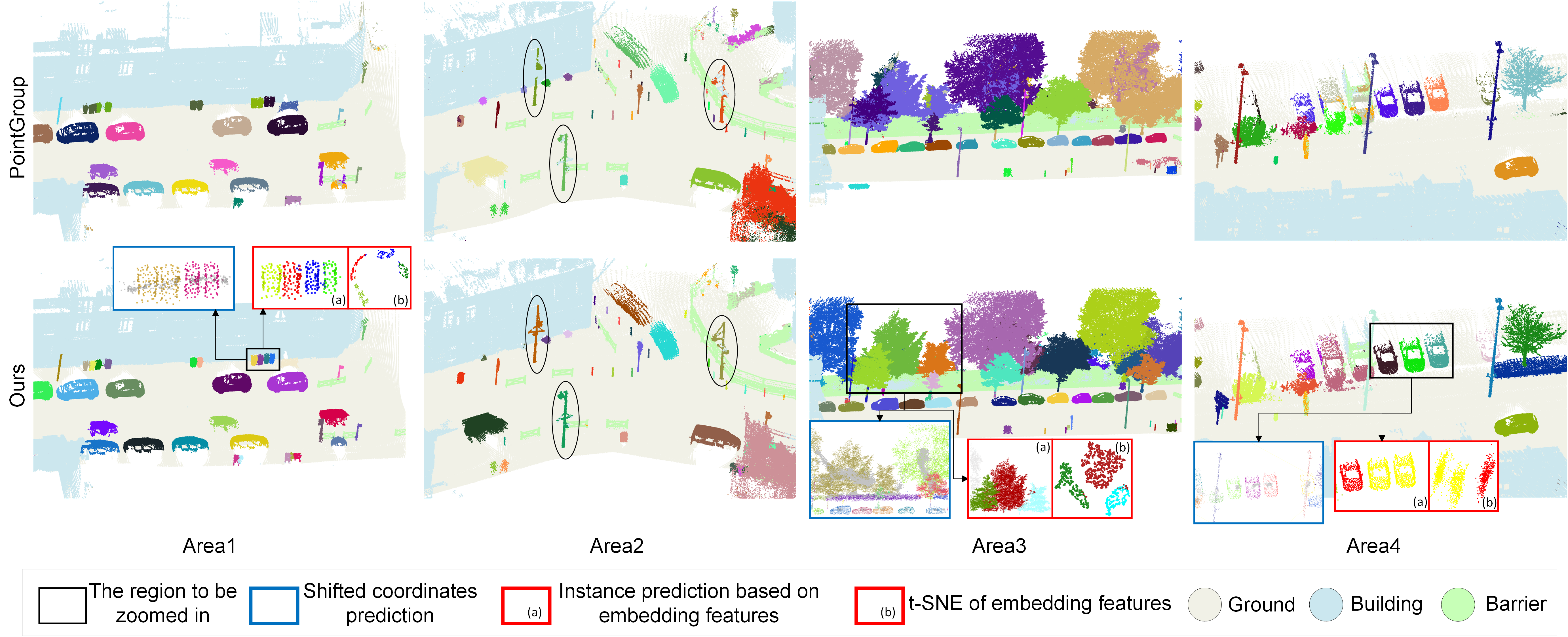}
\vspace{-1.5em}
\caption{Example segmentations from the NPM3D dataset. See section~\ref{sec:EvaluationOnNPM3D} of the text for a discussion of these results. Light gray, light blue and light green colours denote uncountable ``stuff'' classes, saturated random colours denote instances.}
\vspace{0.5em}
\label{fig:QuanlitativeResultsNPM3D}
\end{figure*}

The focus of the present paper is on how to best segment object instances. We compare different designs of the instance clustering branches in Table~\ref{table:5SettingsforInstanceSeg}. \emph{Setting I} corresponds to only the discriminative embedding, without predicting and clustering 3D centroid offsets. Conversely, \emph{setting II} only clusters based on the predicted centroid offsets and does not learn a discriminative embedding. \emph{Setting III} denotes the configuration advocated by PointGroup~\cite{Jiang2020PointGroup}, where the clustering based on centroid offsets is complemented by clustering also the raw 3D points (before shifting them by the offset vectors), and the best instances are selected from the resulting, redundant set of clusters with a ScoreNet. \emph{Setting IV} is the combination of centroid-based clustering and embedding feature clustering (again followed by a ScoreNet), as described above. Finally, \emph{setting V} additionally includes clusters based on raw point coordinates, as advocated by~\cite{Jiang2020PointGroup}, on top of the two cluster sets of setting IV; thus further enlarging the candidate pool, but also making score-based pruning harder. 

All results were computed with four-fold cross-validation: in turn, each sub-regions serves as test set once, whereas the other three are used for training. Then the predictions for all four regions are concatenated to obtain labels for the test dataset, and the performance metrics are calculated.
%\footnote{Note, this is different from averaging the test results of the four regions, since they do not have the same point counts and label distributions.}
The metrics for all five settings are given in Table~\ref{table:5SettingsResultsNpm3d}. It can be seen that the proposed setting IV yields the best balance between precision and recall for instance segmentation (F1-score), as well as the best semantic segmentation (mIoU), and consequently also the highest PQ values. Clustering based solely on either offsets or embedding features significantly reduces precision. The clustering variant introduced by PointGroup, based on raw point coordinates, noticeably reduces recall. It appears that, for quite a number of object instances, the scan point distribution is too diffuse to delineate them. The results for setting V show that instance candidates based on raw points, surprisingly, not only miss many points but even distract from better, competing candidates. It appears that these poorly matching clusters inject noise into the ranking procedure. In other words, complementary methods to diversify the candidate set are only beneficial if the additional candidates are of sufficient quality.

\begin{table*}[tb]
\centering
%\resizebox{\textwidth}{!}{%
\begin{tabular}{c|ccccc|c|ccc|c}
\toprule
\multicolumn{1}{c|}{\multirow{2}{*}{\begin{tabular}[c]{@{}c@{}}Radius\\ (m)\end{tabular}}} &
  \multicolumn{5}{c|}{Instance segmentation [\%]} &
  \multicolumn{1}{c|}{Semantic seg. [\%]} &
  \multicolumn{3}{c}{Panoptic segmentation [\%]} &
  \multicolumn{1}{|c}{\multirow{2}{*}{\begin{tabular}[c]{@{}c@{}}Training\\ time\end{tabular}}} \\ \cmidrule{2-10}
\multicolumn{1}{c|}{} & mCov & mWCov & mPrec & mRec & F1 & mIoU & RQ & SQ & PQ &
  \multicolumn{1}{c}{} \\ \midrule
4 & 65.2 & 78.1 & 58.4 & 65.9 & 61.9 & 96.5 & 81.0 & 88.9 & 73.2 & 3.6 days \\
8 & \underline{68.7} & \underline{81.0} & \underline{69.2} & \underline{68.7} & \textbf{68.9} & \textbf{97.2} & \underline{84.5} & \underline{90.6} & \textbf{77.3} & 7.8 days\\ \bottomrule
\end{tabular}%
%}
\vspace{-0.75em}
\caption{Ablation of cylinder radius for FOR-instance data.}
\vspace{0.25em}
\label{table:radiusAbliationforFOR}
\end{table*}

\begin{table*}[tb]
\centering
%\resizebox{\textwidth}{!}{%
\begin{tabular}{l|ccccc|c|ccc}
\toprule
\multirow{2}{*}{} &
  \multicolumn{5}{c|}{Instance segmentation} &
  \multicolumn{1}{c|}{Semantic seg.} &
  \multicolumn{3}{c}{Panoptic segmentation} \\ \cmidrule{2-10} 
           & mCov  & mWCov & mPrec & mRec  & F1    & mIoU  & RQ    & SQ    & PQ    \\ \midrule
PointGroup & 49.0 & 46.9 & 54.8 & 48.2 & 51.3 & 97.0 & 75.6 & 87.7 & 68.3    \\
Setting IV (5D embed + 3D offset) & \underline{68.7} & \underline{81.0} & \underline{69.2} & \underline{68.7} & \textbf{68.9} & \textbf{97.2} & \underline{84.5} & \underline{90.6} & \textbf{77.3} \\ \bottomrule
\end{tabular}%
%}
\vspace{-0.75em}
\caption{Panoptic segmentation results on FOR-instance data. All values are percentages [\%].}
\vspace{0.25em}
\label{table:PanopticforFOR}
\end{table*}

\begin{figure*}[t!]
\includegraphics[width=1.0\textwidth,trim={0 3mm 0 0},clip]{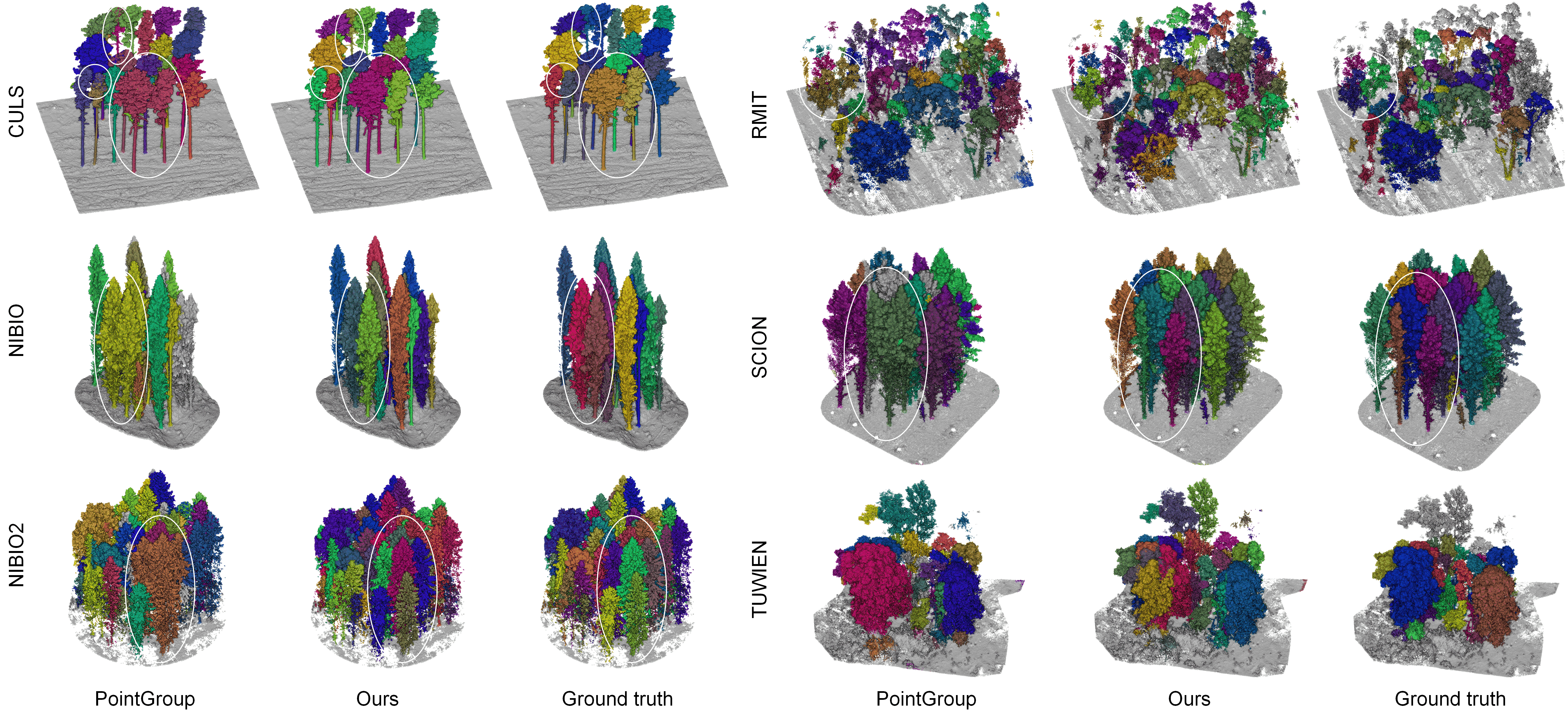}
\vspace{-1.5em}
\caption{Example segmentations from six different regions within the FOR-instance dataset. Gray denotes non-tree points, other colours have been chosen randomly and indicate different instances.}
\label{fig:QuanlitativeResultsFOR}
\end{figure*}

Figure~\ref{fig:QuanlitativeResultsNPM3D} illustrates the differences qualitatively for four representative examples.
In Area 1, adjacent trash cans challenge the instance segmentation. The centroid offset prediction fails to separate them, whereas discriminative embedding succeeds, as can be seen in the t-SNE projection of the 5D embedding space.
Area 2 highlights a case where PointGroup suffers from its hard assertion that only points from the same semantic category can be clustered together. This over-reliance on the category labels means that instance segmentation cannot correct semantic segmentation errors, as on the streetlights marked by a black circle.
Area 3 shows an example for the particularly challenging category of trees, which have large shape variability and are not delimited by well-defined surfaces. When they are located close to each other, the centroid method becomes unreliable, whereas they can still be separated in the discriminative embedding.
Area 4 illustrates the opposite case, where the embedding features are unable to separate two cars, which sometimes occurs especially when there are many instances in close proximity. But since the cars are well enough separated, the centroid offsets are correctly predicted for most of their points and rectify the mistake.

\subsection{Evaluation on FOR-instance}\label{sec:EvaluationOnFORInstance}

The FOR-instance dataset defines a canonical train/test split, to which we adhere. Within the training portion, we randomly set aside 25\% of the data files as our validation set to monitor generalisation and hyper-parameters. As for NPM3D, we concatenate the results of all test sets and compute the performance metrics from that overall segmentation result.

\noindent\textbf{Ablation of voxel size.} Unsurprisingly, the rather simple segmentation between tree and non-tree points is hardly affected by the voxel grid filtering. But also instance segmentation performance is remarkably stable across a wide range of voxel sizes, Figure~\ref{fig:NPM3DAblation}e. It reaches its maximum for voxels with side length 20 to 25$\,$cm, but even at 40$\,$cm the panoptic quality PQ drops less than 2.5 percent points under the maximum. Also very small voxels degrade performance only a little (likely because of diminished spatial context information, due to empty voxels), but significantly increase the training time. From our results, we do not see a reason to decrease the voxel size below 20$\times$20$\times$20$\,$cm$^\text{3}$ for this application. 

\noindent\textbf{Ablation of cylinder radius.} As shown in Table~\ref{table:radiusAbliationforFOR}, expanding the radius of the input blocks from 4$\,$m to 8$\,$m improves all performance metrics. The main reasons is that the bigger radius increases the chance of covering trees completely with a single block, leading to better instance segmentation. For forestry applications we therefore recommend to use rather large neighbourhoods, despite the significantly longer training time.

We also compare our preferred setting, with embedding and offset branches, block radius 8$\,$m and voxel size 20$\,$cm, to our implementation of the PointGroup method, see Table~\ref{table:PanopticforFOR}. We observe a marked improvement of all metrics with our proposed version, with over 17 percent points difference in F1-score. It appears that in the forest setting, where object centroids are hard to estimate and object boundaries are diffuse, the discriminative embedding has a clear advantage over clustering methods that operate in 3D geometric object space.

Figure~\ref{fig:QuanlitativeResultsFOR} shows example results from different locations in the FOR-instance dataset. We note that both tested methods produce surprisingly compelling instance segmentations in most cases, across a range of forest characteristics. Still, our mixed clustering approach consistently yields results on par or better than PointGroup, see differences marked with white ellipses.
FOR-instance was released only recently, and we are not aware of any other published results on the dataset. From the user perspective, we note that our pipeline achieves satisfactory instance segmentation without region-specific parameter tuning or post-processing, challenges commonly reported in the context of tree segmentation, e.g.,~\cite{Wang2021Individual,Chang2022TwoStage,Wilkes2022TLS2trees}.

\section{CONCLUSION}\label{sec:Conclusions}
We have studied the bottom-up approach to panoptic segmentation of outdoor 3D point clouds. We found that the bottleneck is the correct clustering of points into instances, and have constructed a pipeline with two complementary segmentation branches: one that is based on 3D centroid prediction and is well-suited for well-separated, compact objects; and a second one that is based on a discriminative embedding of the 3D points and better handles (nearly) contiguous objects and fuzzy object borders. In experiments on two different datasets, a contemporary panoptic segmentation pipeline with a carefully designed instance clustering stage was able to reach F1-scores \textgreater74\% for objects in an urban mapping context and, remarkably, F1-scores \textgreater68\% for trees in dense forest plots.

{
%\begin{spacing}{1.17}
\begin{spacing}{1.17}
\almostnormal
\bibliography{Bxiang_PanopticSeg}
\end{spacing}
}

\end{document}